\PassOptionsToPackage{square,numbers}{natbib}
\documentclass{article}

\usepackage{arxiv}

\usepackage[utf8]{inputenc} 
\usepackage[T1]{fontenc}    
\usepackage{hyperref}       
\usepackage{url}            
\usepackage{booktabs}       
\usepackage{amsfonts}       
\usepackage{nicefrac}       
\usepackage{microtype}      

\usepackage{graphics} 
\usepackage{epsfig} 
\usepackage{amsmath} 
\usepackage{amssymb}  
\usepackage{amsthm}
\usepackage{cleveref}
\usepackage{subcaption}
\usepackage[numbers]{natbib}

\title{Towards Recurrent Autoregressive Flow Models}

\author{%
  John Mern \\
  Stanford University\\
  Stanford, CA 94305 \\
  \texttt{jmern91@stanford.edu} \\
   \And
   Peter Morales \\
   MIT Lincoln Laboratory \\
   Lexington, MA 02420 \\
   \texttt{peter.morales@mit.ll.edu} \\
   \And
   Mykel J. Kochenderfer \\
   Stanford University \\
   Stanford, CA 94305 \\
   \texttt{mykel@stanford.edu} \\
}

\begin{document}

\maketitle

\begin{abstract}
Stochastic processes generated by non-stationary distributions are difficult to represent with conventional models such as Gaussian processes.
This work presents Recurrent Autoregressive Flows as a method toward general stochastic process modeling with normalizing flows. 
The proposed method defines a conditional distribution for each variable in a sequential process by conditioning the parameters of a normalizing flow with recurrent neural connections. 
Complex conditional relationships are learned through the recurrent network parameters.
In this work, we present an initial design for a recurrent flow cell and a method to train the model to match observed empirical distributions.
We demonstrate the effectiveness of this class of models through a series of experiments in which models are trained on three complex stochastic processes.
We highlight the shortcomings of our current formulation and suggest some potential solutions.


\end{abstract}

\section{Introduction}
Normalizing Flows are the class of invertible transforms with tractably computable Jacobian determinants. 
Normalizing Flow are used to estimate complex probability distributions by transforming a simple base distribution, such as the Gaussian, using the change of variables theorem~\cite{rezende2015}.
These transformed distributions retain many of the attractive features of the base distribution, such as tractable density calculation and efficient sampling, while being flexible enough to model complex, multi-modal densities. 
Because of this, normalizing flows have been used in a variety of learning contexts, particularly variational inference~\cite{kingma2016} and deep generative modeling~\cite{papamakarios2017}.

Existing methods can model complex static distributions but are limited in their ability to model stochastic processes or other complex conditional distributions~\cite{garnelo2018}.
Many normalizing flow models use highly-parameterized function approximators such as neural networks to transform distributions~\cite{dinh2016}. 
Modeling conditional distributions is typically accomplished by changing the parameters of the base distribution, such as the mean of a base Gaussian distribution.
This limits how much the learned transform may be changed by the conditioning term. 

This work seeks to overcome this limitation by introducing a more flexible mechanism for conditioning based on Recurrent Neural Networks (RNNs). 
The output of a RNN at a given time step is a function of the input at that time step and the history of previous inputs in the sequence~\cite{pearlmutter1989}.
This is achieved through use of recurrent cells that maintain hidden states over each step in a sequence. 
In this work, we propose a bijective recurrent cell based on the Gated Recurrent Unit (GRU)~\cite{cho2014}.
We present the Recurrent Autoregressive Flow (RAF), based on this modified GRU, that is able to condition the transform at a given step by the preceding history, allowing for more expressive conditioning than previous methods~\cite{van2016}.

We present the result of several experiments in which RAF graphs were trained to model various complex processes. 
We show that the RAF is generally able to outperform baseline methods on the selected tasks.
The experiments also revealed shortcomings of the existing approach and suggest potential improvements to the current RAF model and training process.

\section{Background}
A normalizing flow is a function $Z = f(X)$ that transforms one random variable to another.
To be a useful flow, it must meet the following criteria. 
\begin{enumerate}
    \item It is invertible, such that $X = f^{-1}(Z)$.
    \item It is a bijection, such that for all $\mathbf{x} \in X$, $f(\mathbf{x})$ maps to only one $\mathbf{z}$ and vice versa.
    \item It has a tractably computable Jacobian determinant $\big|\frac{\partial f(\mathbf{x})}{\partial \mathbf{x}}\big|$.
\end{enumerate}
The probability distribution of a variable $X$ that was transformed from a variable $Z$ through a normalizing flow may be calculated by the change of variables formula as shown in~\cref{eq:CoV}.
\begin{align}
    \label{eq:CoV}
    P_X(\mathbf{x}) &= P_Z(f(\mathbf{x}))\bigg|\text{det}\Big(\frac{\partial f(\mathbf{x})}{\partial \mathbf{x}}\Big)\bigg| \\
    \log \big(P_X(\mathbf{x})\big) &= \log\big(P_Z(f(\mathbf{x}))\big) + \log\bigg|\text{det}\Big(\frac{\partial f(\mathbf{x})}{\partial \mathbf{x}}\Big)\bigg|
\end{align}

Normalizing Flow Graphs (NFGs) are computational graphs composed of layers of simple normalizing flows. 
The probability distribution of variable transformed by a NFG can be calculated by applying the change of variables theorem with the chain-rule. 
This is shown for a NFG with $D$ layers in~\cref{eq:Chain}. 
\begin{align}\label{eq:Chain}
    P_X(\mathbf{x}) &= P_Z(f(\mathbf{x}))\bigg|\text{det}\Big(\prod_{d=0}^D\frac{\partial f^d(\mathbf{x})}{\partial \mathbf{x}^d}\Big)\bigg| \\
    P_X(\mathbf{x}) &= P_Z(f(\mathbf{x}))\bigg|\prod_{d=0}^D\text{det}\Big(\frac{\partial f^d(\mathbf{x})}{\partial \mathbf{x}^d}\Big)\bigg| 
\end{align}

Samples from $P_X$ are generated by sampling $\mathbf{z} \sim P_Z$ and then transforming $\mathbf{z}$ by the inverse of the NFG as $\mathbf{x} = f^{-1}(\mathbf{z})$. 
In general, calculating the determinant of the Jacobian $\frac{\partial f(\mathbf{x})}{\partial \mathbf{x}}$ is $\mathcal{O}(d^3)$ for a $d \times d$ matrix. 
In order to tractably compute the density of $\mathbf{x}$, normalizing flows are designed such that the Jacobian takes a special form with an easy to calculate determinant. 

\section{Prior Work}
Normalizing flows have been proposed in various forms for variational inference and generative modeling. 
A popular early model was the planar flow model, in which variables were transformed through a series of simple additive operations~\cite{rezende2015}. 
Nonlinear Independent Components Estimation (NICE) uses additive coupling layers similar to planar flows, alternating with multiplicative rescaling layers~\cite{dinh2014}.
In NICE models, the input is first partitioned into two parts $\mathbf{x} \in \mathbb{R}^D \rightarrow \mathbf{x}_{1:d}, \ \mathbf{x}_{d:D}$. 
The output $\mathbf{y}\in\mathbb{R}^D$ is calculated from each part separately such that $\mathbf{y}_{1:d} = \mathbf{x}_{1:d}$ and $\mathbf{y}_{d:D} = \mathbf{x}_{d:D} + m(\mathbf{x}_{1:d})$. 
The Jacobian of a NICE transform is then a triangular matrix.

Both Planar Flows and NICE flows are volume-preserving, meaning they have a Jacobian determinant of $1$.
While these models can translate the density of a distribution, they cannot scale the variance. 
This limits their ability to map variables onto distributions of much smaller or larger support volume than the base distribution. 

In order to overcome this limitation, RealNVP~\cite{dinh2016} flows were proposed as an extension to NICE.
RealNVP flows partition and shift the input just as in NICE, adding a scale term such that $\mathbf{y}_{d:D} = \mathbf{x}_{d:D}\odot exp(g(\mathbf{x}_{1:d})) + m(\mathbf{x}_{1:d})$.
MADE flows also use a similar approach~\cite{germain2015}.
With this scale factor, the Jacobian is lower-triangular and the determinant is no longer necessarily $1$. 
The transforms of the RealNVP and MADE models can be represented as 
\begin{align}\label{eq:partition}
    & \mathbf{z} = \mathbf{x} \odot f(\mathbf{x}_{0:k}) + g(\mathbf{x}_{0:k}) \\
    & \mathbf{x} \in \mathbb{R}^K, \ k<K 
\end{align}
Decomposing the input $\mathbf{x}$ to $\mathbf{x}_{0:k}, \mathbf{x}_{k:K}$, we can write the Jacobian as
\begin{align}\label{eq:Triangular}
    \frac{\partial \mathbf{z}}{\partial \mathbf{x}} & =
    \begin{bmatrix}
    \text{diag}(\textbf{a}) & \textbf{0} \\
    \frac{\partial z_{k:K}}{\partial x_{0:k}} & \text{diag}(\textbf{b})
    \end{bmatrix} \\
    \textbf{a} = f(\mathbf{x}_{0:k})_{0:k} & + \mathbf{x}_{0:k} \odot f'(\mathbf{x}_{0:k}), \ \textbf{b} = f(\mathbf{x}_{0:k})_{k:K}
\end{align}
This Jacobian is lower triangular, and the determinant can be easily calculated as the trace of the matrix in $\mathcal{O}(d)$ time.

Inverse Autoregressive Flows (IAFs) introduced the concept of auto-regressive models as normalizing flows~\cite{kingma2016}. 
These models make the assumption that sample ordering is sequential, such that $P(\mathbf{x}_i|\mathbf{x}_{0:i-1})$. 
The IAF flow transform is shown in~\cref{eq:IAF}. 
The Masked Autoregressive Flow (MAF)~\cite{papamakarios2017} uses the same transform as IAF but makes the sequential assumption on the latent variables $Z$.
Both IAF and MAF have a Jacobian of the form given in~\cref{eq:Triangular}. 
\begin{align}\label{eq:IAF}
    & \mathbf{z}_i = \mathbf{x}_i \odot \sigma(\mathbf{x}_{0:i-1}) + \mu(\mathbf{x}_{0:i-1}) 
\end{align}
Despite the name, these autoregressive models were not specifically developed for sequential data. 
In many cases, flows operate along the sample dimension such that for a sample $\mathbf{x}\in\mathbb{R}^D$, any given element $\mathbf{x}_i$ can be conditionally dependent only on elements $\mathbf{x}_{j<i}$. 
Each sample, however, is treated as independently drawn.
It has been empirically shown that, even with this assumption, autoregressive flows can be effective generative models~\cite{van2016}.
Commonly, autoregressive flow graphs alternate autoregressive layers with permutation layers to reduce the effect of the ordering assumption. 
By changing the variable order at each transform, the order dependence across the entire graph can be reduced. 
Permutations are themselves normalizing flows.

All of the above transforms use a simple Hadamard-product and affine transforms at each layer. 
The Neural Autoregressive Flow (NAF)~\cite{huang2018} model uses a feed-forward neural network as a transform layer. 
By composing the network with strictly positive weights and strictly monotonic activation functions, the resulting transform is invertible. 
Unfortunately, as pointed out in the Block NAF extension~\cite{DeCao2019}, there is no easy analytic method to invert the neural network transform.
As a result, a given NAF model can only be used for inference or generative sampling, not both as is the case with other models. 

Some work has been done to develop flows to directly model random sequences. 
Latent normalizing flows~\cite{ziegler2019} used autoregressive flows with a latent state variable to represent sequential processes. 
These flows, however, are limited to sequences of discrete variables. 
\section{Methods}
Existing auto-regressive flow methods do not directly address temporally correlated sequences.
Some methods have introduced conditioning vectors to in the transform such that $\mathbf{z}_t = f(\mathbf{x}_t, \phi(\mathbf{x}_{t-1}))$.
These methods typically use simple affine functions to condition the transforms, limiting how expressive the dynamic models can be.
They are also limited by only including the effect of the single prior time step as a conditioning signal. 
The current work extends these approaches to allow conditioning on an arbitrary length history of previous steps. 

One mechanism well suited to encoding temporal data into compact representations is the Recurrent Neural Network. 
RNNs learn temporal relations by propagating signals forward in time through cell hidden states. 
These hidden states carry information from one time step to the next and also introduce a path by which training gradients may be back propagated. 
In this way, RNNs can learn optimal embeddings of sequences as hidden-states, which can be viewed as a form of short-term memory. 

\begin{figure}
    \centering
    \includegraphics[width=0.7 \columnwidth]{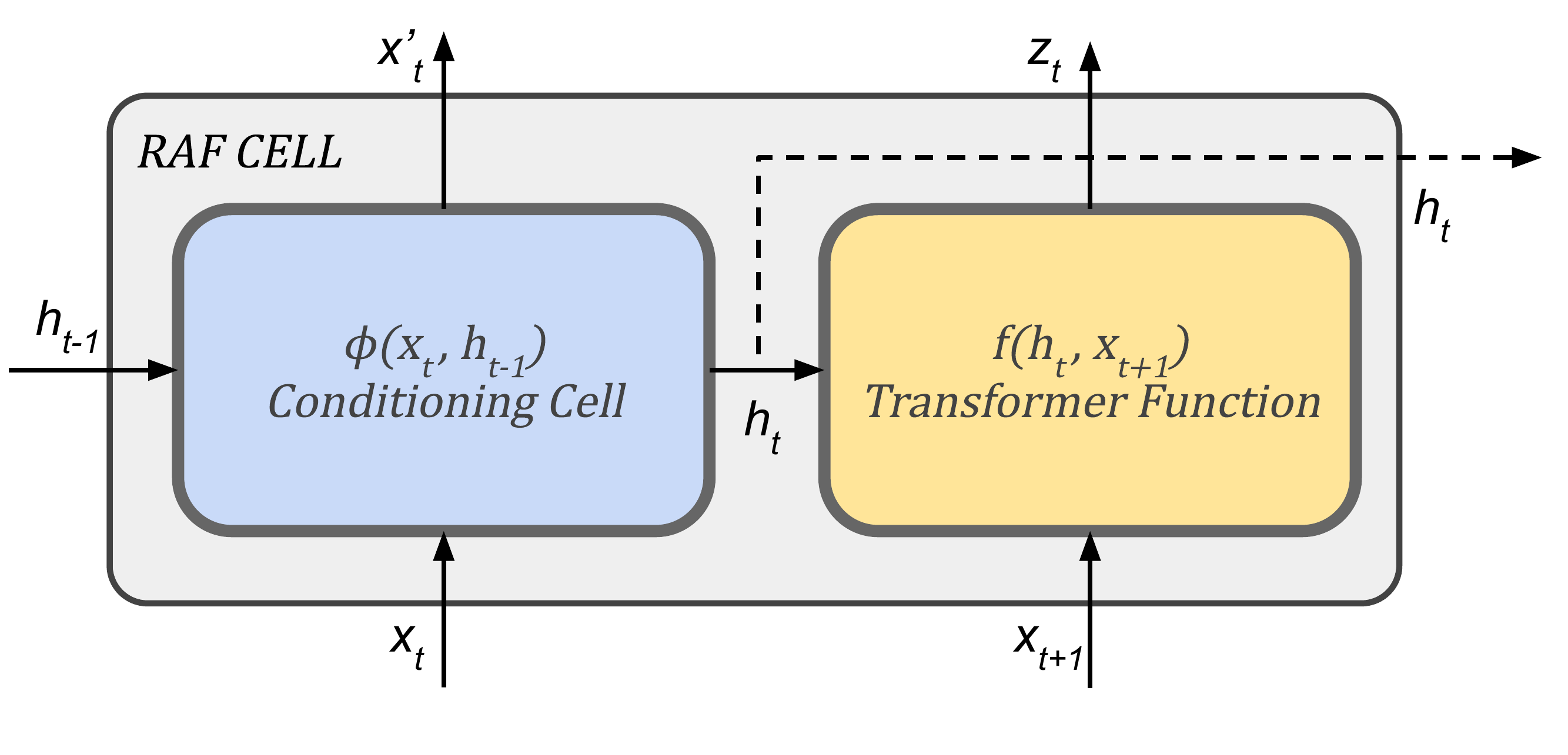}
    \caption{Recurrent normalizing flow cell. The cell state $h_t$ evolves as a function of the observed input $x_t$ at each time step and the previous cell state $h_{t-1}$. The transformer function outputs the transformed variable value $z_{t+1}$ given the input $x_{t+1}$, conditioned on the hidden state value. The transform may be in forward $z_{t+1} = f(x_{t+1})$ or inverse $x_{t+1} = f^{-1}(z_{t+1}, h_t)$. This figure shows the cell in the inverse configuration.}
    \label{fig:GRU}
\end{figure}

We incorporated this ability by developing a normalizing flow with a recurrent connection as shown in~\cref{fig:GRU}. 
Similar to the IAF and MAF, we represent our transforms as conditioning functions coupled transformer functions. 
The recurrent conditioning function is based on the Gated Recurrent Unit (GRU) which uses gate functions to control the evolution of the hidden state in time.
The GRU was selected because of its tendency to stabilize gradients in the temporal direction during training~\cite{cho2014}.
At each time step, the Normalizing Flow Graph (NFG) models the distribution of the future state $P(\mathbf{x}_{t+1}\mid \mathbf{x}_{0:t})$. 
Each recurrent layer of the NFG transforms the predicted variable $\hat{\mathbf{x}}_{t+1}$ using the transformer function conditioned on the cell hidden state $h_t$. 

Each layer takes as input the hidden state of the previous time stepas well as the current sample.
Each layer outputs the cell hidden state and either the $\mathbf{z}_t$ or $\hat{\mathbf{x}}_{t+1}$ depending if it is operating in forward or inverse mode. 
When generating samples of $\hat{\mathbf{x}}_t$ from samples of $\mathbf{z}_t$, the NFG is operating in forward mode. 
When estimating density of a given sample $\mathbf{x}_t$ by transforming it to $\mathbf{z}_t$, the NFG is operating in inverse mode.

The hidden-state update is shown in~\cref{eq:GRU}. 
\begin{align}\label{eq:GRU}
    \mathbf{y}_t &= \sigma(W_y \mathbf{x}_t + U_y \mathbf{h}_{t-1} + \mathbf{b}_y)\\
    \mathbf{r}_t &= \sigma(W_r \mathbf{x}_t + U_r \mathbf{h}_{t-1} + \mathbf{b}_r)\\
    \mathbf{h}_t &= (1 - \mathbf{y}_t) \odot \mathbf{h}_{t-1} + \mathbf{y}_t \odot \tanh(W_h \mathbf{x}_t + U_h(\textbf{r}_t \odot \mathbf{h}_{t-1}) + \mathbf{b}_h)  
\end{align}
The RAF cell can use any bijective flow as a transformer function. 
The proposed transformer function first splits the hidden state vector into a weight vector $\mathbf{h}^{(w)}_t$ and bias vector $\mathbf{h}^{(b)}_t$. 
The weight vector is reshaped into a lower-triangular matrix $W^{(h)}_t$.
An affine transform is applied with the weight matrix and bias vector, and an invertable non-linear function $\sigma$ is applied to the resulting value. 
The complete transform is 
\begin{align}\label{eq:sigma flow}
    \mathbf{\mathbf{z}}_t = \sigma(W^{(h)}_t\mathbf{x}_t + \mathbf{h}^{(b)}_t)  
\end{align}

The conditioning signal is only a function of $\mathbf{x}_{t-1}$ for the transform of $\mathbf{x}_t$, so the Jacobian of the transform with respect to $\mathbf{x}_t$ is still lower triangular.
This enables complex temporal dependencies to be learned without increasing computational complexity of the change of variables operations. 
Additionally, the use of a triangular matrix for the affine transform enables the learning of more complex relationships between the values of each forward sample $\mathbf{x}_{t+1}$ than do previously proposed affine transforms.
Like the element-wise affine transform, the triangular matrix does imply some ordering dependency that may not exist in the real data. 
In this way, these transforms are autoregressive and some permutation layers are still required. 

We can derive the Jacobian of the recurrent flow by first applying the chain rule and finding the Jacobians of the affine and non-linearity transforms. 
For non-linearity transforms (such as sigmoid), the value of each output dimension $z_t^i$ is a function only of the corresponding input value $x^i$, so the Jacobian is a diagonal matrix. 
For the affine transform, the determinant is the weight matrix $W_h$.
The complete Jacobian can be calculated as
\begin{align}
    z_t = \sigma(&g(x_{t+1})), \ g(x_{t+1}) = W_h x_t \\
    \frac{\partial z_t}{\partial x_{t+1}} &= \frac{\partial z_t}{\partial g(x_{t+1})} \frac{\partial g(x_{t+1})}{\partial x_{t+1}} \\
    \frac{\partial z_t}{\partial g(x_{t+1})} &= \text{diag}\big[\sigma(g(x_{t+1}))(1 - \sigma(g(x_{t+1})))\big] \\
    \frac{\partial g(x_{t+1})}{\partial x_{t+1}} &= W_h
\end{align}
for a given non-linear activation function $\sigma$.
We can then find the determinants of each term independently as the product of the diagonal values as
\begin{align}\label{eq:determinant}
    \text{det}\Big(\frac{\partial z_t}{\partial g(x_{t+1})}\Big) &= \prod_{i=0}^K \sigma(g(x_{t+1}^i))(1 - \sigma(g(x_{t+1}^i))) \\
    \log\bigg|\text{det}\Big(\frac{\partial z_t}{\partial g(x_{t+1})}\Big)\bigg| &= \sum_{i=0}^K \log\big|\sigma(g(x_{t+1}^i))(1 - \sigma(g(x_{t+1}^i)))\big| \\
    \text{det}\Big(\frac{\partial g(x_{t+1})}{\partial x_{t+1}}\Big) &= \prod_{i=0}^K W_t^{ii} \\
    \log\bigg|\text{det}\Big(\frac{\partial g(x_{t+1})}{\partial x_{t+1}}\Big)\bigg| &= \sum_{i=0}^K \log\big|W_t^{ii}\big| 
\end{align}
where the $ii$ indexes the diagonal elements of the weight matrix. 
Unlike NAFs, RAF inverse transforms can be easily calculated analytically as the product of the inverse triangular matrices. 


\section{Experiments}
We ran three experiments testing the performance of RAF graphs in modeling stochastic processes with various challenging features.
The process in the first experiment is a hierarchical stochastic process in 2D space. 
In the second experiment, the RAF predicts the trajectory of a stochastic agent performing a simple navigation task. 
In the third experiment, the RAF models the evolution of a fluid flow represented as a non-stationary bimodal distribution.

\subsection{Experiment 1: Hierarchical Stochastic Process}
Previous flow methods have been able to model complex stationary distributions through topology transforms.
This experiment compares the performance of an existing flow method to RAF in modeling a non-stationary distribution generated through a simple stochastic process. 

\begin{figure}[h!]
    \centering
    \begin{subfigure}[t]{0.32\textwidth}
        \centering
        \includegraphics[height=1.4in]{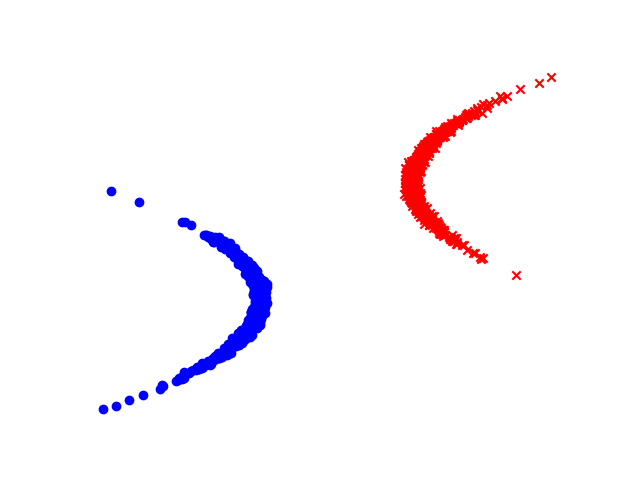}
        \caption{True Distribution}
    \end{subfigure}%
    ~ 
    \begin{subfigure}[t]{0.32\textwidth}
        \centering
        \includegraphics[height=1.4in]{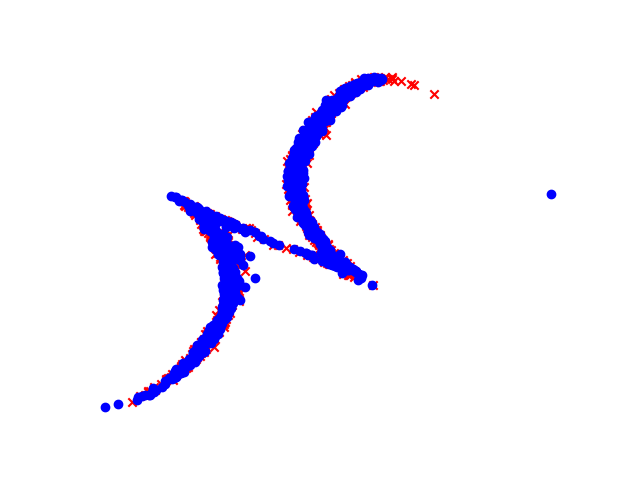}
        \caption{Real-NVP Model}
    \end{subfigure}
    ~
     \begin{subfigure}[t]{0.32\textwidth}
        \centering
        \includegraphics[height=1.4in]{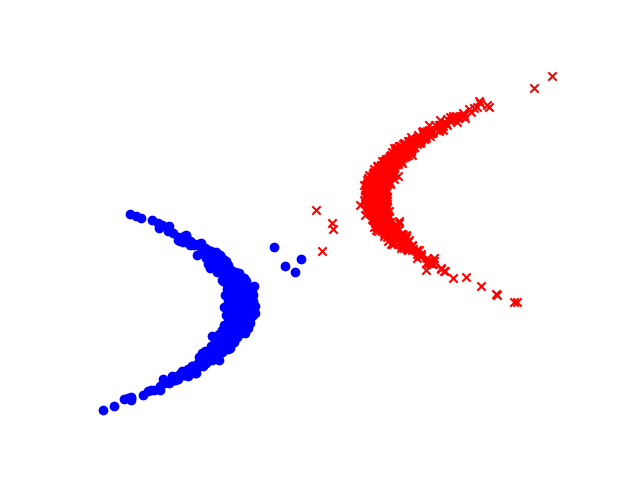}
        \caption{RAF Model}
    \end{subfigure}
    \caption{Hierarchical process modeling results. Each subplot shows data generated in one episode with $y^{(i)} = 0$ in blue and $y^{(i)} = 1$ in red. 
    }
    \label{fig:Exp1}
\end{figure}

The process generates data in batches or episodes. 
At the start of each episode, the process draws an index from a Bernoulli distribution $y^{(i)} \sim \text{Bern}(p)$.
The observed data points $X^{(i)}$ are then drawn sampling from Gaussian distributions as
\begin{align}
    &x^{(i)}_{j1} \sim \mathcal{N}(\mu^{(i)}, 4\mathbb{I}) \\
    &x^{(i)}_{j0} \sim \mathcal{N}(0.25\mu^{(i)}(x^{(i)}_{j1})^2, \mathbb{I}) 
\end{align}
where $\mu^{(i)} = 2y^{(i)} - 1$.
The resulting distribution is shown in~\cref{fig:Exp1}.

We trained a Real-NVP and an RAF to model this process. 
Both graphs had five layers and transformed a Standard-Normal distribution. 
They were each trained with 10,000 samples generated across 100 episodes. 
The training loss was the negative log-likelihood of the data. 
The Adam optimizer~\cite{Kingma2014} was used with an initial learning rate of $10^{-3}$.

After training, each model was run for ten episodes of 1,000 samples each. 
As can be seen in~\cref{fig:Exp1}, while the Real-NVP approximately recovers the shape of the distribution of the training data, samples from a single generative batch are generated in both the $y=0$ and $y=1$ regions. 
A single episode from the true process would contain points from only a single mode. 
This suggests the Real-NVP was unable to learn the hierarchical generative process.
Further, the Real-NVP graph was unable to completely separate the base normal distribution into the two components of the original distribution using. 
The static topology transform left a manifold connecting the two modes. 

For the RAF, we visualized the data and selected one episode for each $y^{(i)}$ value. 
As with the Real-NVP, the RAF accurately captures the shape of the training data distribution. 
In addition, it was able to learn the underlying generative process. 
Each episode contains samples only from a single mode of the distribution. 
A few points are generated between the distributions at the start of each episode.
This is likely due to the model starting with zero hidden-state and accumulating hidden state upon observing samples. 
For generative modeling, this can be overcome with a short burn-in period. 

\subsection{Experiment 2: Stochastic Trajectory Prediction}
In the second experiment, we trained a RAF model to predict the position of an agent completing a simple navigation task. 
In the task, a mouse navigates a maze to the cheese at the goal position.
At each time step, the mouse agent moves toward the goal along the corridor in which it is currently located.
Each step it moves a distance of 1 unit with probability 0.8 or 2 units with probability 0.2. 
When the mouse reaches corridor intersection, it chooses to change directions with probability 0.5.
This procedure is a hierarchical stochastic process operating on varying timescales, with the agent speed sampled every time step and the direction sampled only at intersection instances.

We trained an RAF graph to predict the distribution of the next state of the mouse given the previously observed trajectory $P(x_{t+1}\mid x_{0:t})$. 
The graph had 8 layers of RAFs of 64 hidden units each. 
The model was trained to maximize the average log probability density of observed positions across each episode. 
The Adam optimizer was used with initial learning rate of $10^{-3}$ over 10,000 epochs. 

As a baseline, we also trained an RNN to output parameters for a multi-variate Gaussian distribution over the next state.
\begin{align}
    (\mu_t, \tau_t) &= \text{RNN}(x_{0:t}) \\
    x_{t+1} &\sim \mathcal{N}(\mu_t, \tau_t\mathbb{I})
\end{align}

After training, we ran the RAF and RNN each for 50 episodes and measured the average log-probability density of the predictions at each time step.
The RAF had an average log probability density of $1.83 \pm 0.31$. 
The RNN had an average log probability density of $1.2 \pm 0.12$.

\begin{figure}[h]
    \centering
    \includegraphics[width=0.95\textwidth]{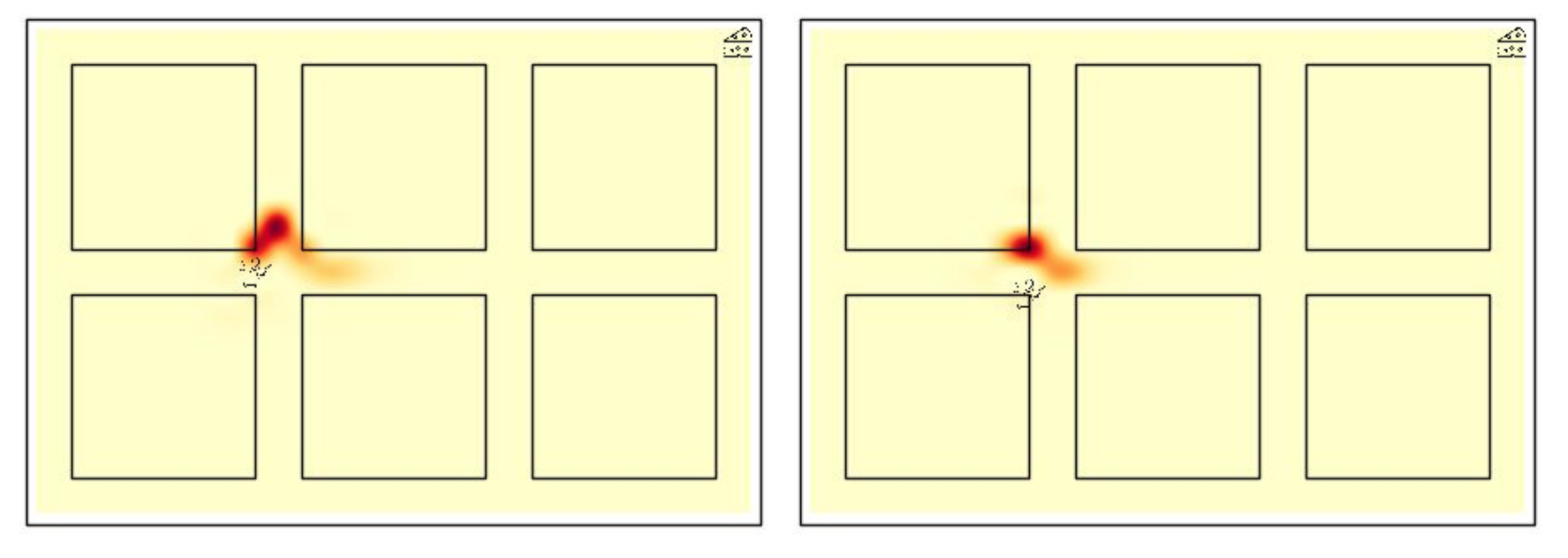}
    \caption{Visualization of the maze navigation agent experiment. The two images show instances where the agent approached the same corridor intersection traveling from the left to the right. As can be seen, the RAF only partially maps the distribution to the correct shape, with different results at each approach.}
    \label{fig:Exp2a}
\end{figure}

Qualitative inspection of the test episodes showed that the RAF was able to learn to predict the bimodal position distribution caused by the speed distribution.
However, the graph was only partly successful in learning multi-modal distributions at the corridor intersections. 
\Cref{fig:Exp2a} shows two instances in which the agent has entered the same intersection travelling from left to right. 
We would expect approximately equal probability mass to be assigned to continuing to travel right as to traveling vertically. 
We see, however, that in both cases significantly more mass is assigned to the vertical travel direction, and that significant probability mass is assigned to unreachable positions. 
This shortcoming is likely due to the difficulties associated with RNN predictions over multiple time-scales, as the frequency of intersection occurrences is far lower than the base time step frequency. 
Only two modes were observed in the distributions at the intersections, where three were expected. 
This may suggest the fixed transform parameters learned by the model were more tuned to modeling the bimodal speed distribution and not adequately affected by the conditioning signal.
Incorporating importance sampling into the training process to better account for sparse events may be used to overcome this deficiency. 

\subsection{Experiment 3: Fluid flow modeling}
For our final experiment, we trained a RAF to model a simple fluid-flow.
This was selected to test the ability of the RAF to model non-stationary continuous distributions under unknown dynamics.

\begin{figure}
    \centering
    \includegraphics[width=0.22\textwidth]{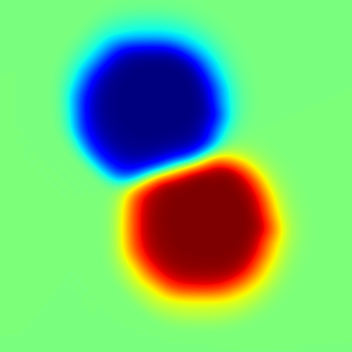}
    \includegraphics[width=0.22\textwidth]{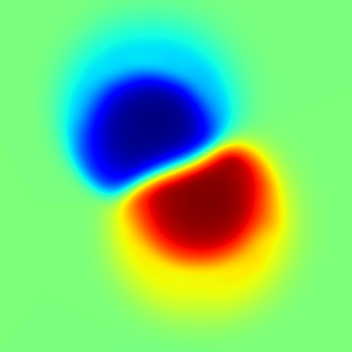}
    \includegraphics[width=0.22\textwidth]{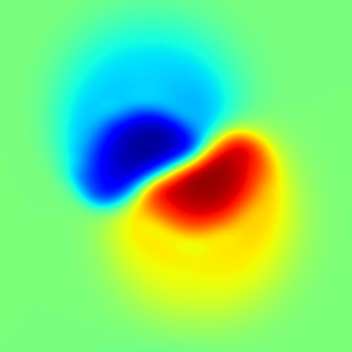}
    \includegraphics[width=0.22\textwidth]{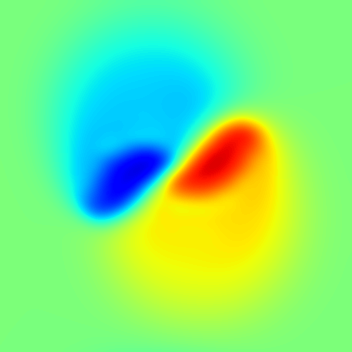}
    \subcaptionbox*{t=0}
    {\includegraphics[width=0.22\textwidth]{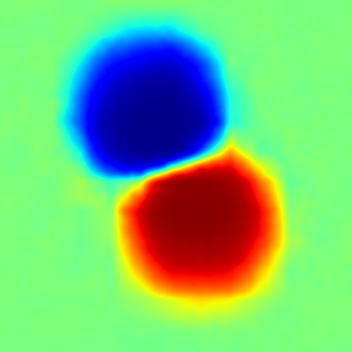}}
    \subcaptionbox*{t=2}
    {\includegraphics[width=0.22\textwidth]{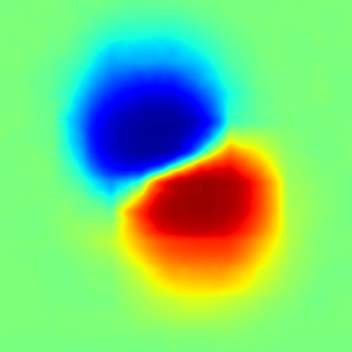}}
    \subcaptionbox*{t=4}
    {\includegraphics[width=0.22\textwidth]{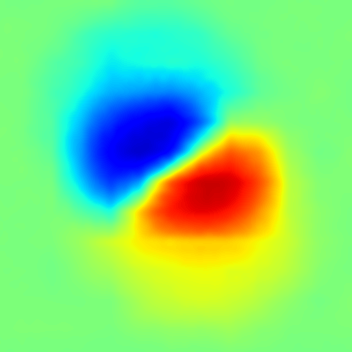}}
    \subcaptionbox*{t=6}
    {\includegraphics[width=0.22\textwidth]{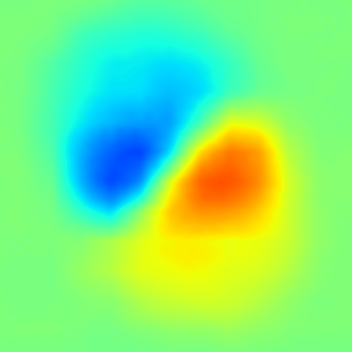}}
    \caption{Visualization of the flow modeling task. The top row shows the true fluid motion evolving over multiple time steps. The bottom row shows the predicted fluid motion from the RAF. The RAF is able to accurately model this physical phenomenon by casting it to a distribution modeling challenge, with some accuracy loss.}
    \label{fig:flow}
\end{figure}

For this test, we built a 2D flow simulator using simplified linear heat and mass transfer equations. 
In each episode a mass of hot fluid is initialized below the vertical mid-plane of the environment and offset from the horizontal mid-plane by a random distance sampled from a zero-mean Gaussian distribution. 
A second cold fluid mass is initialized the same way above the mid plane. 
During the episode, the cold and hot masses sink and rise respectively at rates approximately proportional to their temperatures.
Heat is also transferred to neighboring fluid, causing temperatures to tend toward equal. 
Snapshots from an episode are shown in the top row of~\cref{fig:flow}.

In order to train an RAF to model this, we created a 2D distribution from each simulation step. 
To do this, we numerically integrated exponential of the fluid temperature $e^{t_x}$ over the flow area and normalized the resulting quantity such that the resulting field would be a proper distribution.
No heat was allowed to exit the control-volume, so the normalization constant remained fixed throughout the episode.
The resulting distribution is difficult to fit since it not only has a strong maximum at the point of high-temperature, but also a strong minimum at the low-temperature point, with the surrounding area having moderate probability density. 

At each step, our RAF predicted the next distribution $P(x_{t+1}\mid o_{0:t})$, where $x_t$ is a position in the plane and $o_t$ is the observation of the distribution at time $t$.
To train this, we minimized the average KL-divergence between the predicted distribution $RAF(x_{0:t}) = Q(x_{t+1}\mid o_{0:t};\theta)$ and the true distribution. 
We estimated the KL divergence through a discrete 2D approximation as 
\begin{equation}
    D_{KL}^t(\theta) = \sum_i\sum_j P(x_{i,j}\mid o_{0:t}) \frac{\log P(x_{i,j}\mid o_{0:t})}{\log Q(x_{i,j}\mid o_{0:t};\theta)}  
\end{equation}
where $x{i,j}$ is the value of the field at point $i,j$.

An autoencoder was pre-trained to vectorize the 2D field to allow it to be used by the RAF graph. 
An autoencoder is a neural network that encodes an input to a compressed latent space and decodes the latent representation to an estimate of the original input, typically with some data loss. 
The latent state of the autoencoder used was a 32-dimensional vector. 
The autoencoder was trained to minimize empirical L1 loss on the reconstruction as
\begin{equation}
    L1(\phi) = \frac{1}{m}\sum_{k=0}^m\sum_{(i,j)}|\hat{x}_{(i,j)} - x_{i,j}|
\end{equation}
where $m$ is the number of samples in a batch of training inputs, $x_{i,j}$ is the true field value and $\hat{x}_{i,j}$ is the reconstructed value.

The autoencoder had five convolutional layers each for the encoder and decoder, with kernel size $3 \times 3$, stride 2, and a $\tanh$ activation functions. 
The flow graph had 16 layers with hidden size 128.
The base distribution was a standard Normal distribution. 

The end-to-end network was trained on a composite loss 
\begin{equation}
    L(\theta, \phi) = \text{L1}(\phi) + \alpha D_{KL}(\theta, \phi)
\end{equation}
where $\alpha$ is a hyper-parameter weighting KL-divergence loss relative to reconstruction loss.
The network was trained for 3,000 epochs using the Adam optimizer. 

After training, the RAF was run for 100 episodes for evaluation. 
The average log probability density of each step was $72.22 \pm 0.16$. 
The predicted field and actual field for an example episode are shown for predictions at times 0, 2, 4, and 6 in~\cref{fig:flow}.

As can be seen in the figure, the flow graph is able to produce samples that are perceptually similar to the original fluid simulation, despite the challenging multi-modal dynamics.
This suggests that having dense data available over the whole problem domain for each time step aided the training process. 
There is, however, significant blur on the modes at the later time steps. 
Some of this blur is likely a result of the loss from encoding the 2D field to the vector representation for RAF. 

\section{Conclusion}
This work introduced Recurrent Autoregressive Flows (RAFs) as a new class of normalizing flow model.
Through the introduction of a recurrent connection, RAFs allow flow models to learn more complex temporal relationships over sequential data than previous conditional flow methods. 
In this work, we showed that our proposed transform meets the required characteristics of a normalizing flow in that it is analytically invertible, bijective, and has an efficiently computable Jacobian determinant. 

Our experiments show that RAFs improve performance of existing flow methods in modeling stochastic processes. 
Future work will study how to better compose the conditioning signal of the RAF with more complex flow layers, such as the block neural auto-regressive flow.
Work will also be done to better condition on events occurring at varying time-scales. 
One approach that will be considered is a temporal dilation approach as in the clockwork RNN~\cite{koutnik2014}.
Another approach to be considered will be the use of an importance-weighted replay buffer to enable learning of rare stochastic events~\cite{mnih2016}. 

\subsubsection*{Acknowledgments}
DISTRIBUTION STATEMENT A. Approved for public release. Distribution is unlimited.

This material is based upon work supported by the Under Secretary of Defense for Research and Engineering under Air Force Contract No. FA8702-15-D-0001. Any opinions, findings, conclusions or recommendations expressed in this material are those of the author(s) and do not necessarily reflect the views of the Under Secretary of Defense for Research and Engineering.

© 2020 Massachusetts Institute of Technology.

Delivered to the U.S. Government with Unlimited Rights, as defined in DFARS Part 252.227-7013 or 7014 (Feb 2014). Notwithstanding any copyright notice, U.S. Government rights in this work are defined by DFARS 252.227-7013 or DFARS 252.227-7014 as detailed above. Use of this work other than as specifically authorized by the U.S. Government may violate any copyrights that exist in this work.

The authors acknowledge the MIT SuperCloud and Lincoln Laboratory Supercomputing Center for providing (HPC, database, consultation) resources that have contributed to the research results reported within this paper/report.

\bibliographystyle{IEEEtran}
\bibliography{main.bib}
\end{document}